# Merging Knowledge Bases in Possibilistic Logic by Lexicographic Aggregation

**Guilin Qi**
Southeast University
Nanjing 211189, China

**Jianfeng Du**
Guangdong University of Foreign Studies
Guangzhou 510006, China

**Weiru Liu, David Bell**
Queen's University Belfast
Belfast, BT7 1NN, UK

## Abstract

Belief merging is an important but difficult problem in Artificial Intelligence, especially when sources of information are pervaded with uncertainty. Many merging operators have been proposed to deal with this problem in possibilistic logic, a weighted logic which is powerful for handling inconsistency and dealing with uncertainty. They often result in a *possibilistic knowledge base* which is a set of weighted formulas. Although possibilistic logic is inconsistency tolerant, it suffers from the well-known "drowning effect". Therefore, we may still want to obtain a consistent possibilistic knowledge base as the result of merging. In such a case, we argue that it is not always necessary to keep weighted information after merging. In this paper, we define a merging operator that maps a set of possibilistic knowledge bases and a formula representing the integrity constraints to a classical knowledge base by using lexicographic ordering. We show that it satisfies nine postulates that generalize basic postulates for propositional merging given in [11]. These postulates capture the principle of minimal change in some sense. We then provide an algorithm for generating the resulting knowledge base of our merging operator. Finally, we discuss the compatibility of our merging operator with propositional merging and establish the advantage of our merging operator over existing semantic merging operators in the propositional case.

## 1 Introduction

Logic-based belief merging is an important topic in Artificial Intelligence and has application in many computer science fields, such as multi-agent systems and requirements engineering. In a logical framework, each information source is often considered as a knowledge base, which is a set of logical formulas. Even though each knowledge base is consistent, putting them together may give rise to logical contradiction. Since an inconsistent knowledge base is useless due to the fact "ex falso quodlibet" (a false statement implies an arbitrary statement), we need to resolve inconsistency when we merge knowledge bases. Furthermore, in practice, information is often pervaded with uncertainty. This is because the truth of some propositions may remain unknown in the presence of incomplete or partial information.

Possibilistic logic [6] provides a flexible framework to handle inconsistency and deal with uncertainty. At the syntactic level, it is a weighted logic which attaches to each formula with a weight belonging to a totally ordered scale, such as $[0,1]$, where the weight is interpreted as the certainty level of the formula. A *possibilistic knowledge base* is a set of weighted formulas. At the semantic level, it is based on the notion of a *possibility distribution*, which is a mapping from the set of interpretations $\Omega$ to interval $[0,1]$. For each possibilistic knowledge base, there is a unique possibility distribution associated with it.

Many merging approaches have been proposed in possibilistic logic [3, 4, 2, 5, 17, 14, 16, 12]. In [4, 2], given several possibilistic knowledge bases, a semantic combination rule (or merging operator) is applied to aggregate the possibility distributions associated with them into a new possibility distribution. Then the syntactical counterpart of the semantic merging of the possibility distributions is a possibilistic knowledge base whose associated possibility distribution is the new possibility distribution [2]. Several approaches (such as those in [17]) have been proposed to improve the semantic combination rules. The approach proposed in [1] is based on argumentation. The model-based approach given in [14] is based on a model-based merging opera-tor in propositional logic. There are other approaches for merging possibilistic knowledge bases that are either based on some context information [12] or based on a negotiation framework [16].

Almost all merging approaches in possibilistic logic result in a possibilistic knowledge base. Although possibilistic inference is inconsistency-tolerant, it suffers from the notorious "drowning effect", and so we may still want to obtain a consistent merged possibilistic knowledge base (see [17, 12, 16]). In such a case, it is not always necessary to keep weighted information in the resulting knowledge base after merging. For example, suppose we are interested in knowing if a formula can be inferred from the consistent merged knowledge base and do not really need to consider what degree it can be inferred, the weighted information is not useful any more. This is because a formula can be inferred from a consistent possibilistic knowledge base through possibilistic inference if and only if it can be inferred from the classical knowledge base associated with the possibilistic knowledge base, where a classical knowledge base is one with non-weighted information. There are some disadvantages for over-emphasizing the pursuit of keeping weighted information in the result of merging. First, it is difficult to define a possibilistic merging operator which is compatible with propositional merge. When applied to merge propositional knowledge bases directly[1](we attach weight 1 to every formula in the knowledge bases), a possibilistic merging operator usually does not satisfy all the basic postulates for propositional merging given in [3] or in [11] (see [13] for a discussion). Furthermore, regarding possibilistic merging preserving weighted information, as far as we known, there does not exist a set of postulates in possibilistic logic that generalize postulates in [10] or [11] such that an existing semantic merging operator[2] satisfies all of them. The absence of such a set of postulates makes it hard to evaluate the rationality of a semantic merging operator in possibilistic logic.

In this paper, we propose a merging operator in possibilistic logic that maps a set of possibilistic knowledge bases and a formula representing the integrity constraints to a classical knowledge base. The models of the result of our merging operator are models of the formula representing the integrity constraints that are maximal w.r.t. the lexicographic ordering. Therefore, the result of merging of our operator is syntactically a propositional knowledge base. Although lexicographic ordering has been used to define some important merging operators in propositional logic, its application to possibilistic merging is still unexplored. We generalize the set of basic postulates for propositional merging given in [11] and show that our merging operator satisfies all of them. These postulates capture the principle of minimal change in some sense. We then provide an algorithm for generating the resulting knowledge base of our merging operator. Finally, we discuss the compatibility of our merging operator with propositional merging. We show that our merging operator can be reduced to a propositional merging operator that satisfies all the basic postulates given in [11]. According to the relationship between possibilistic logic and ordinal conditional functions, it is not difficult to see that our merging operator can be also used to merging uncertain information in ordinal conditional functions.

The rest of this paper proceeds as follows. We give some preliminaries on possibilistic logic in next section. We then define our merging operator in possibilistic logic and discussion its logical properties. In the next section, the syntactic counterpart of our model-based merging operator is given. After that, we discuss related work. Finally we conclude the paper and present some future work.

## 2 Preliminaries on Possibilistic Logic

We consider a propositional language $\mathcal{L}_{PS}$ over a finite set $PS$ of propositional symbols. The classical consequence relation is denoted as $\vdash$. An interpretation is a total function from $PS$ to $\{true, false\}$. $\Omega$ is the set of all possible interpretations. An interpretation $\omega$ is a model of a formula $\phi$ iff $\omega(\phi) = true$. For each formula $\phi$, we use $Mod(\phi)$ to denote its set of models. A *(classical) knowledge base* $B$ is a finite set of propositional formulas which can be identified with the conjunction of its elements. $K$ is consistent iff $Mod(K) \neq \emptyset$. Two knowledge bases $K_1$ and $K_2$ are equivalent, denoted $K_1 \equiv K_2$, iff $Mod(K_1) = Mod(K_2)$. A *knowledge profile* $E$ is a multi-set of knowledge bases, i.e. it may contain a knowledge base twice. The union of multi-sets will be denoted by $\sqcup$.

The semantics of possibilistic logic [6] is based on the notion of a *possibility distribution* $\pi$ which is a mapping from the set of interpretations $\Omega$ to interval [0,1]. The possibility degree $\pi(\omega)$ represents the degree of compatibility (resp. satisfaction) of the interpretation $\omega$ with the available beliefs about the real world. From a *possibility distribution* $\pi$, two measures can be determined: the possibility degree of formula $\phi$, $\Pi_\pi(\phi) = max\{\pi(\omega) \mid \omega \in \Omega, \omega \models \phi\}$ and the necessity degree of formula $\phi$, $N_\pi(\phi) = 1 - \Pi_\pi(\neg \phi)$.

At the syntactic level, a formula, called a *possibilis-*

---

[1] It has been shown that some well-known distance-based merging operators can be encoded by possibilistic merging in [2]. This is not a direct application of possibilistic merging to propositional knowledge bases.

[2] By a semantic merging operator we mean a merging operator that is semantically defined by possibility distributions.

*tic formula*, is represented by a pair $(\phi, a)$, where $\phi$ is a propositional formula and $a$ is an element of the semi-open real interval $(0, 1]$ or of a finite total ordered scale, which means that the necessity degree of $\phi$ is at least equal to $a$, i.e. $N(\phi) \geq a$. Then uncertain or prioritized pieces of information can be represented by a *possibilistic knowledge base* which is a finite set of possibilistic formulas of the form $B = \{(\phi_i, a_i) : i = 1, ..., n\}$. The classical base associated with $B$ is denoted as $B^*$, namely $B^* = \{\phi_i \mid (\phi_i, a_i) \in B\}$. A possibilistic knowledge base $B$ is consistent if and only if its classical base $B^*$ is consistent. A *possibilistic knowledge profile* $\mathcal{E}$ is a multi-set of possibilistic knowledge bases.

Given a possibilistic knowledge base $B$, a unique *possibility distribution*, denoted by $\pi_B$, can be obtained by the principle of minimum specificity [6]. For all $\omega \in \Omega$,

$$\pi_B(\omega) = \begin{cases} 1 & \text{if } \forall(\phi_i, a_i) \in B, \omega \models \phi_i, \\ 1 - max\{a_i \mid \omega \not\models \phi_i, (\phi_i, a_i) \in B\} & \text{otherwise.} \end{cases} \quad (1)$$

**Definition 1** *Let $B$ be a possibilistic knowledge base, and $a \in [0, 1]$. The $a$-cut (resp. strict $a$-cut) of $B$ is $B_{\geq a} = \{\phi_i \in B^* \mid (\phi_i, b_i) \in B \text{ and } b_i \geq a\}$ (resp. $B_{>a} = \{\phi_i \in B^* \mid (\phi_i, b_i) \in B \text{ and } b_i > a\}$).*

**Definition 2** *Let $B$ be a possibilistic knowledge base. The* inconsistency degree *of $B$ is:*

$$Inc(B) = max\{a_i \mid B_{\geq a_i} \text{ is inconsistent}\},$$

with $Inc(B) = 0$ if $B$ is consistent. The inconsistency degree of $B$ is the largest weight $a_i$ such that the $a_i$-cut of $B$ is inconsistent.

**Definition 3** *Let $B$ be a possibilistic base. A formula $\phi$ is said to be a consequence of $B$ to a degree $a$, denoted by $B \vdash_\pi (\phi, a)$, iff (i)$B_{\geq a}$ is consistent; (ii) $B_{\geq a} \vdash \phi$; (iii)$\forall b > a, B_{\geq b} \not\vdash \phi$.*

Two possibilistic knowledge bases $B$ and $B'$ are said to be equivalent, denoted by $B \equiv_s B'$, iff $\forall\ a \in (0, 1]$, $B_{\geq a} \equiv B'_{\geq a}$. Two possibilistic knowledge profiles $\mathcal{E}_1$ and $\mathcal{E}_2$ are said to be equivalent ($\mathcal{E}_1 \equiv_s \mathcal{E}_2$) iff there is a bijection between them such that each possibilistic knowledge base of $\mathcal{E}_1$ is equivalent to its image in $\mathcal{E}_2$.

## 3 A Merging Operator in Possibilistic Logic by Lexicographic Aggregation

Lexicographic ordering has been used to define merging operators in propositional logic [11, 7] and stratified knowledge bases [15]. It has been shown that these merging operators are rational in the sense that they satisfy the set of basic postulates and some other additional postulates for propositional merging given in [11] or the generalized set of basic postulates. However, its application to define a merging operator in possibilistic logic is unexplored. In this section, we define our merging operator in possibilistic logic by the lexicographic ordering and show that it satisfies some interesting logical postulates.

### 3.1 Definition

We recall the notion of lexicographical ordering on vectors of numbers (see [8] for more details).

**Definition 4** *Given two vectors of numbers $\vec{a} = (a_1, ..., a_n)$ and $\vec{b} = (b_1, ..., b_n)$. Let $\sigma$ and $\sigma'$ be two permutations on $\{1, ..., n\}$ such that $a_{\sigma(i)} \geq a_{\sigma(i+1)}$ and $b_{\sigma'(i)} \geq b_{\sigma'(i+1)}$ for all $i$. The lexicographical ordering $\leq_{lex}$ between $\vec{a}$ and $\vec{b}$ is defined as:*

$\vec{a} \leq_{lex} \vec{b}$ *if and only if $a_{\sigma(i)} = b_{\sigma'(i)}$ for all $i$ or there exists $i \geq 1$ such that $a_{\sigma(i)} < b_{\sigma'(i)}$ and $a_{\sigma(j)} = b_{\sigma'(j)}$ for all $1 \leq j < i$.*

The lexicographical ordering compares two sequences of numbers in a descending order (we use two permutations to assure that the sequences of numbers to be compared are in descending order). As usual, $\vec{a} <_{lex} \vec{b}$ denotes $\vec{a} \leq_{lex} \vec{b}$ and $\vec{b} \not\leq_{lex} \vec{a}$, and $\vec{a} \equiv_{lex} \vec{b}$ denotes $\vec{a} \leq_{lex} \vec{b}$ and $\vec{b} \leq_{lex} \vec{a}$.

We define our merging operators using the lexicographical ordering.

**Definition 5** *Let $\mathcal{E} = \{B_1, ..., B_n\}$ be a possibilistic knowledge profile, and $\mu$ be a formula representing the integrity constraint. Suppose $\pi_i$ is the possibility distribution associated with $B_i$, for each $i = 1, ..., n$. For each interpretation $\omega$, we can associate with it a vector of numbers $\vec{l}_\mathcal{E}(\omega) = (\pi_1(\omega), ..., \pi_n(\omega))$. The resulting knowledge base of lexicographical ordering based merging operator, denoted by $\Delta_\mu^{Lex}(E)$, is defined in a model-theoretic way as follows:*

$\omega \in Mod(\Delta_\mu^{Lex}(\mathcal{E}))$ *if and only if $\omega \in Mod(\mu)$ and $\forall \omega' \in Mod(\mu), \vec{l}_\mathcal{E}(\omega) \geq_{lex} \vec{l}_\mathcal{E}(\omega')$.*

In Definition 5, each interpretation is associated with a vector consisting of its possibility degrees relative to all possibilistic knowledge bases. Then any two interpretations can be compared $w.r.t$ the lexicographical ordering defined by Definition 4. The models of the resulting knowledge base of operator $\Delta^{Lex}$ are the models of $\mu$ that are maximal $w.r.t$ the lexicographical ordering. The result of our merging operator is a propositional knowledge base, instead of a possibilistic knowledge base. The advantages of our merging operator over existing merging operators in possibilistic logic are that our merging operator fulfils the principle of minimal change and is compatible with propositional merging, as we will show later. As we have argued in the Introduction, suppose we want to resolve inconsistency after merging, it is not always necessary to keep weighted information in the resulting knowledge base of merging. However, it is not desirable to use our merging operator for iterated merging. This is because our merging operator drops the weighted information after merging, which is critical for possibilistic merging. In order to iterate the merging process, after merging a set of possibilistic knowledge bases, we may have to keep the original knowledge bases in case further information is received.

| $\omega$ | value | $B_1$ | $B_2$ | $B_3$ | $B_4$ | $\vec{l}_{\mathcal{E}}(\omega)$ |
|---|---|---|---|---|---|---|
| $\omega_1$ | 0111 | 0.4 | 1.0 | 1.0 | 0.1 | (0.4,1.0,1.0,0.1) |
| $\omega_2$ | 0110 | 0.4 | 1.0 | 1.0 | 0.1 | (0.4,1.0,1.0,0.1) |
| $\omega_3$ | 0011 | 0.1 | 0.4 | 0.4 | 0.1 | (0.1,0.1,0.4,0.1) |
| $\omega_4$ | 0010 | 0.1 | 0.4 | 0.4 | 0.1 | (0.1,0.1,0.4,0.1) |
| **$\omega_5$** | **1111** | **1.0** | **0.4** | **1.0** | **0.4** | **(1.0,0.4,1.0,0.4)** |
| **$\omega_6$** | **1110** | **1.0** | **0.4** | **1.0** | **0.4** | **(1.0,0.4,1.0,0.4)** |

Table 1: $\Delta^{Lex}$ operator

**Example 1** *Let $\mathcal{E} = \{B_1, B_2, B_3, B_4\}$ be a possibilistic profile consisting of four possibilistic knowledge bases, where*

- $B_1 = \{(p_1 \vee p_2, 0.9), (p_3, 0.9), (p_1, 0.6), (p_2, 0.6)\}$;
- $B_2 = \{(p_3 \vee p_4, 0.9), (\neg p_1, 0.6), (p_2, 0.6)\}$;
- $B_3 = \{(p_3, 0.9), (p_2, 0.6)\}$;
- $B_4 = \{(p_1, 0.9), (p_2, 0.8), (\neg p_3, 0.6)\}$.

*In addition, we have an integrity constraint $\mu = (\neg p_1 \vee p_2) \wedge p_3$. The computations of possibility degrees of models of $\mu$ are given in Table 1.*

*In Table 1, the column corresponding to $B_i$ gives the possibility degree of $\omega$ w.r.t. $\pi_i$. The column corresponding to $\vec{l}_{\mathcal{E}}(\omega)$ gives the lists of possibility degrees of interpretations. By Definition 5, it is easy to see that $\omega_5$ and $\omega_6$ are two interpretations with highest priority w.r.t. the lexicographic ordering in Table 1. So $Mod(\Delta_\mu^{Lex}(\mathcal{E})) = \{1111, 1110\}$. Therefore, we have $\Delta_\mu^{Lex}(\mathcal{E}) \equiv p_1 \wedge p_2 \wedge p_3$.*

### 3.2 Logical properties

The following postulates generalize postulates (IC0)-(IC3) and (IC5)-(IC8) for propositional merging given in [11] in a straightforward way. Let $\mathcal{E}, \mathcal{E}_1, \mathcal{E}_2$ be possibilistic knowledge profiles, and $\mu, \mu_1, \mu_2$ be formulas from $\mathcal{L}_{PS}$. Let $\Delta$ be a merging operator that maps a set of possibilistic knowledge bases to a propositional knowledge base.

**(P1)** $\Delta_\mu(\mathcal{E}) \vdash \mu$.

**(P2)** If $\mu$ is consistent, then $\Delta_\mu(\mathcal{E})$ is consistent.

**(P3)** Let $\bigwedge \mathcal{E} = \wedge_{B_i \in \mathcal{E}} \wedge_{\phi_{ij} \in B_i} \phi_{ij}$. If $\bigwedge \mathcal{E}$ is consistent with $\mu$, then $\Delta_\mu(\mathcal{E}) \equiv \bigwedge \mathcal{E} \wedge \mu$.

**(P4)** If $\mathcal{E}_1 \equiv_s \mathcal{E}_2$ and $\mu_1 \equiv \mu_2$, then $\Delta_{\mu_1}(\mathcal{E}_1) \equiv \Delta_{\mu_2}(\mathcal{E}_2)$.

**(P5)** $\Delta_\mu(\mathcal{E}_1) \wedge \Delta_\mu(\mathcal{E}_2) \vdash \Delta_\mu(\mathcal{E}_1 \sqcup \mathcal{E}_2)$

**(P6)** If $\Delta_\mu(\mathcal{E}_1) \wedge \Delta_\mu(\mathcal{E}_2)$ is consistent, then $\Delta_\mu(\mathcal{E}_1 \sqcup \mathcal{E}_2) \vdash \Delta_\mu(\mathcal{E}_1) \wedge \Delta_\mu(\mathcal{E}_2)$

**(P7)** $\Delta_{\mu_1}(\mathcal{E}) \wedge \mu_2 \vdash \Delta_{\mu_1 \wedge \mu_2}(\mathcal{E})$

**(P8)** If $\Delta_{\mu_1}(\mathcal{E}) \wedge \mu_2$ is consistent, then $\Delta_{\mu_1 \wedge \mu_2}(\mathcal{E}) \vdash \Delta_{\mu_1}(\mathcal{E}) \wedge \mu_2$

(P1) requires the result of merging satisfying the integrity constraints. (P2) states that the result of merging should be consistent if the integrity constraints are consistent. (P3) says that if there is not contradiction between the integrity constraints and the knowledge bases, then we keep all the information. (P4) is the condition for syntax-irrelevance. (P5) and (P6) together state that if there exist two subgroups which agree on at least one alternative, then the result of the global merging will be exactly those alternatives the two groups agree on. (P7) and (P8) together say that the notion of closeness is well-behaved, i.e., that an alternative that is preferred among the possible alternatives $\mu_1$, will remain preferred if one restricts the possible choices $\mu_1 \wedge \mu_2$. Thus (P7) and (P8) are important to ensure minimal change.

Next we consider the generalization of (IC4) in [11] given as follows: If $\phi \vdash \mu$ and $\phi' \vdash \mu$, then $\Delta(\{\phi\} \sqcup \{\phi'\}) \wedge \phi \not\vdash \bot$ implies that $\Delta(\{\phi\} \sqcup \{\phi'\}) \wedge \phi' \not\vdash \bot$, where $\phi, \phi'$ and $\mu$ are propositional formulas. This is not trivial because the condition part of the postulate needs to be modified to take into account the priority information. Our new postulate is inspired from a postulate given in [5]. We first introduce the notion of a priority degree of a subbase which was originally defined in [5].

**Definition 6** *Let $B'$ be a possibilistic knowledge base. We define its priority degree w.r.t. another possibilistic knowledge base $B$, denoted by $\mathrm{Deg}_B(B')$, by*

$$\mathrm{Deg}_B(B') = min\{a \mid (\phi, a) \in B' \cap B\},$$

*and $\mathrm{Deg}_B(B') = 1$ if $B \cap B'$ is empty.*

$\mathrm{Deg}_B(B')$ is the priority level of the least certain formulas in $B'$ which also appear in $B$.

Given two possibilistic knowledge bases $B_1$ and $B_2$ and a formula $\mu$, we say that a subset $C$ of $B_1 \cup B_2$ is a *conflict set w.r.t.* $\mu$ if it satisfies (1) $C^* \cup \{\mu\}$ is inconsistent and (2) for any subset $C' \subset C$, $C' \cup \{\mu\}$ is consistent. Two possibilistic knowledge bases $B_1$ and $B_2$ are said to be *equally prioritized w.r.t* $\mu$ if for each conflict set $C$ of $B_1 \cup B_2$ w.r.t. $\mu$, we have $\mathsf{Deg}_{B_1}(C) = \mathsf{Deg}_{B_2}(C)$.

We propose (P9) to generalize (IC4).

**(P9)** Let $B_1$ and $B_2$ be two consistent possibilistic knowledge bases. If $B_1 \vdash_\pi (\mu, a)$ and $B_2 \vdash_\pi (\mu, a)$, and $B_1$ and $B_2$ are equally prioritized w.r.t. $\mu$, then $\Delta_\mu(\{B_1, B_2\}) \wedge \bigwedge_{\phi \in B_1^*} \phi$ is consistent if and only if $\Delta_\mu(\{B_1, B_2\}) \wedge \bigwedge_{\psi \in B_2^*} \psi$ is consistent.

It says that given two consistent possibilistic knowledge bases $B_1$ and $B_2$, if both $B_1$ and $B_2$ are consistent with $\mu$, and they are equally prioritized with respect to $\mu$, then the result of merging is consistent with $B_1$ if and only if it is consistent with $B_2$.

Given a possibilistic profile $\mathcal{E} = \{B_1, ..., B_n\}$ and a formula $\mu$, let $Inc_\mu(\mathcal{E}) = Inc(B_1 \cup ... \cup B_n \cup \{(\mu, 1)\})$, which we call the inconsistency degree of $\mathcal{E}$ w.r.t. $\mu$, we give another postulate for possibilistic merging:

**(P10)** For any possibilistic profile $\mathcal{E} = \{B_1, ..., B_n\}$ and any formula $\mu$, we have $\Delta_\mu(\mathcal{E}) \vdash (\phi, a)$ for any $(\phi, a) \in B_1 \cup ... \cup B_n$ where $a > Inc_\mu(\mathcal{E})$.

(P10) says that the result of a merging operator should keep those possibilistic formulas whose weights are greater than the inconsistent degree of $\mathcal{E}$ w.r.t. $\mu$.

We show that our merging operator satisfies all the generalized postulates.

**Proposition 1** *Merging operator $\Delta^{Lex}$ satisfies (P1)-(P10).*

### 3.3 Computational complexity

Let $\Delta$ be a merging operator. The following decision problem is denoted as MERGE($\Delta$):

- **Input :** a triple $\langle \mathcal{E}, \mu, \psi \rangle$ where $\mathcal{E} = \{B_1, ..., B_n\}$ is possibilistic profile, $\mu$ is a formula, and $\psi$ is a formula.

- **Question :** Does $\Delta_\mu(\mathcal{E}) \models \psi$ hold?

We can establish the following complexity result.

**Proposition 2** *MERGE($\Delta^{Lex}$) is $\Delta_2^p$-complete.*

## 4 Syntactical Counterpart of Our Merging Operator

We propose Algorithm 1 to compute the resulting knowledge base of our merging operator. We use $\Phi$ to denote the set of pairs consisting of a formula $\phi_S$ and a

---

**Algorithm 1**: Algorithm for syntactical computation

**Data**: a possibilistic knowledge profile $\mathcal{E} = \{B_1, ..., B_n\}$; a formula $\mu$
**Result**: a new knowledge base $B$

1 **begin**
2     Let $\Phi := \{(\mu, \mathcal{E})\}$; $Inc := 0$;
3     **while** $\exists (\phi_S, \mathcal{E}_S) \in \Phi, \mathcal{E}_S \neq \emptyset$ **do**
4       $maxc := -\infty$;
5       **foreach** $(\phi_S, \mathcal{E}_S) \in \Phi$ **do**
6         **foreach** $B_j \in \mathcal{E}_S$ **do**
7           Compute $Inc(\phi_S, B_j)$;
8         $Inc_S := min_{B_j \in \mathcal{E}_S} Inc(\phi_S, B_j)$;
9       $Inc := min_{(\phi_S, \mathcal{E}_S) \in \Phi} Inc_S$;
10      $\Phi := \Phi \setminus \{(\phi_S, \mathcal{E}_S) \in \Phi \mid Inc_S \neq Inc\}$; $\Phi' := \emptyset$;
11      **foreach** $(\phi_S, \mathcal{E}_S) \in \Phi$ **do**
12        $I := \{j \mid Inc(\phi_S, B_j) = Inc, B_j \in \mathcal{E}_S\}$;
13        $\mathsf{MCS}(I) := \{J \subseteq I \mid \bigwedge_{j \in J}(B_j)_{>Inc} \wedge \phi_S \not\vdash \bot$ and $\forall k \in I \setminus J, \bigwedge_{j \in J \cup \{k\}}(B_j)_{>Inc} \wedge \phi_S \vdash \bot\}$;
14        $maxc_S := max_{J' \in \mathsf{MCS}(I)} |J'|$;
15        $\mathsf{CardM}(I) := \{J \in \mathsf{MCS}(I) \mid |J| = maxc_S\}$;
16        **foreach** $J \in \mathsf{CardM}(I)$ **do**
17          $\phi_J := \bigwedge_{j \in J}(B_j)_{>Inc} \wedge \phi_S$;
          $\mathcal{E}_J := \mathcal{E}_S \setminus \{B_j \in \mathcal{E}_S \mid j \in J\}$;
18        **if** $maxc_S > maxc$ **then**
19          $\Phi' := \{(\phi_J, \mathcal{E}_J) \mid J \in \mathsf{CardM}(I)\}$
20          $maxc := maxc_S$
21        **else if** $maxc_S = maxc$ **then**
22          $\Phi' = \Phi' \cup \{(\phi_J, \mathcal{E}_J) \mid J \in \mathsf{CardM}(I)\}$;
23      $\Phi := \Phi'$;
24    $B = \vee_{(\phi_S, \mathcal{E}_S) \in \Phi} \phi_S$;
25    **return** $B$
26 **end**

---

knowledge profile $\mathcal{E}_S$, where $\phi_S$ is obtained by merging some selected knowledge bases from $\mathcal{E}$ and $\mathcal{E}_S$ contains knowledge bases which are left to be merged under the integrity constraints $\phi_S$. Given a possibilistic knowledge base $B$ and a formula $\phi$, we use $Inc(\phi, B)$ to denote the inconsistency degree $Inc(B')$ of the new knowledge base $B' = B \cup \{(\phi, 1)\}$ and call it the inconsistency degree of $\phi$ w.r.t. $B$. Initially, $\Phi$ contains a single element $(\mu, \mathcal{E})$. In the "while" step, we check whether there is a pair $(\phi_S, \mathcal{E}_S)$ in $\Phi$ such that $\mathcal{E}_S$ is nonempty. If not, then the algorithm stops and the result of merging is returned (see Proposition 4), which is the disjunction of $\phi_S$ for all $(\phi_S, \mathcal{E}_S)$ where $\mathcal{E}_S$ is nonempty. Otherwise, for each element $(\phi_S, \mathcal{E}_S)$ in $\Phi$, we compute the inconsistency degree $Inc_S$ of $\phi_S$ w.r.t. $\mathcal{E}_S$ as the minimum of $Inc(\phi_S, B_j)$ for all $B_j \in \mathcal{E}_S$ and let $Inc$ be the minimal inconsistency degree among all $Inc_S$, which we call the global inconsistency degree of $\Phi$ (lines 5-9). This global inconsistency degree corresponds to the possibility degree of a model of the result of merging w.r.t. one of the possibilistic knowledge bases. Formally, we have the following proposition.

**Proposition 3** *Let $\mathcal{E} = \{B_1, ..., B_n\}$ be a possibilistic knowledge profile, and $\mu$ be a formula representing the integrity constraint. For any model $\omega$ of $\Delta_\mu^{Lex}(\mathcal{E})$, let $\vec{l}_\mathcal{E}(\omega) = (\pi_1(\omega), ..., \pi_n(\omega))$. Suppose $a_1 > a_2 > ... > a_m$ are all the distinct possibility degrees appearing in $\vec{l}_\mathcal{E}(\omega)$. If Inc is obtained during the k-th iteration in the while loop, then $Inc = 1 - a_k$.*

We then delete those pairs $(\phi_{S'}, \mathcal{E}_{S'})$ such that $Inc_{S'}$ is not equal to the global inconsistency degree from $\Phi$ (line 10). For each such pair $(\phi_{S'}, \mathcal{E}_{S'})$, the possibility degrees of the models of $\phi_{S'}$ must be less than $a_k$ given in Proposition 3 if Inc is obtained during the k-th iteration in the while loop. That is, none of the models of $\phi_{S'}$ can be maximal w.r.t the lexicographical ordering, thus we do not need to handle $(\phi_{S'}, \mathcal{E}_{S'})$. According to Proposition 3, $Inc = 1 - a_k$ for some $a_k$ in $\vec{l}_\mathcal{E}(\omega)$ where $\omega$ is a model of $\Delta_\mu^{Lex}(\mathcal{E})$. For each $(\phi_S, \mathcal{E}_S)$ left in $\Phi$, we need to obtain the maximum $maxc_S$ of the cardinalities of subsets $T$ of $\mathcal{E}_S$ such that the set of Inc-cut sets of possibilistic knowledge bases in $T$ is consistent with $\phi_S$ (line 14). We use $maxc$ to obtain the maximum of all $maxc_S$ associated with $(\phi_S, \mathcal{E}_S)$ in $\Phi$. For those pairs $(\phi_S, \mathcal{E}_S)$ in $\Phi$ such that $maxc_S \neq maxc$, all models $\omega$ of $\phi_S$ cannot be maximal w.r.t. the lexicographical ordering because the number of possibilistic knowledge bases $B_j$ in $\mathcal{E}_S$ such that $\pi_{B_j}(\omega) = 1 - Inc$ must be less than $maxc$ (remember that $1 - Inc = a_k$ if Inc is obtained during the k-th iteration in the while loop by Proposition 3). Thus these pairs are dropped (lines 18-22). For pairs $(\phi_S, \mathcal{E}_S)$ in $\Phi$ such that $maxc_S = maxc$, for each cardinally maximal subset $T$ of $\mathcal{E}_S$ such that the set of Inc-cut sets of possibilistic knowledge bases in $T$ is consistent with $\phi_S$, we obtain a formula $\phi_J$ which is the conjunction of $\phi_S$ and the Inc-cut sets of possibilistic knowledge bases in the subset are consistent with $\phi_S$. A set $\mathcal{E}_J$, which is the complement of this cardinally maximal subset by $\mathcal{E}_S$, is then attached to the new formula $\phi_J$ for further merging (lines 16-17). $\Phi$ is reset to contain all those pairs of $(\phi_J, \mathcal{E}_J)$ (line 23) and we go back to the "while" step again.

**Example 2** *(Continue Example 1)* Initially, we have $\Phi = \{(\mu = (\neg p_1 \vee p_2) \wedge p_3, \mathcal{E} = \{B_1, B_2, B_3, B_4\})\}$. We have $Inc(\mu, B_1) = 0$, $Inc(\mu, B_2) = 0$, $Inc(\mu, B_3) = 0$ and $Inc(\mu, B_4) = 0.6$. So $Inc = 0$. There is only one element in $\Phi$, so $\Phi$ is not changed. Since $Inc(\mu, B_1) = Inc(\mu, B_2) = Inc(\mu, B_3) = 0$, we have $I = \{1, 2, 3\}$. It is not difficult to get that $\mathsf{MCS}(I) = \mathsf{CardM}(I) = \{\{1,3\}, \{2,3\}\}$ and $maxc_\emptyset = 2$. So $\phi_{\{1,3\}} = p_1 \wedge p_2 \wedge p_3$ and $\mathcal{E}_{\{1,3\}} = \{B_2, B_4\}$, and $\phi_{\{2,3\}} = \neg p_1 \wedge p_2 \wedge p_3$ and $\mathcal{E}_{\{2,3\}} = \{B_1, B_4\}$. Since $maxc_\emptyset > maxc = -\infty$, we have $\Phi' = \{(\phi_{\{1,3\}}, \mathcal{E}_{\{1,3\}}), (\phi_{\{2,3\}}, \mathcal{E}_{\{2,3\}})\}$ and $\Phi = \Phi'$.

We go to the second iteration of the while loop. Note $maxc$ is reset to $-\infty$. We have $Inc(\phi_{\{1,3\}}, B_2) = 0.6$, $Inc(\phi_{\{1,3\}}, B_4) = 0.6$, $Inc(\phi_{\{2,3\}}, B_1) = 0.6$ and $Inc(\phi_{\{2,3\}}, B_4) = 0.9$. Thus $Inc_{\{1,3\}} = Inc_{\{2,3\}} = 0.6$. So $Inc = 0.6$ and $\Phi$ is not changed. We handle $(\phi_{\{1,3\}}, \mathcal{E}_{\{1,3\}})$ first. We have $I = \{2, 4\}$. It is not difficult to check that $\mathsf{MCS}(I) = \mathsf{CardM}(I) = \{I\}$ and $maxc_{\{1,3\}} = 2$. We have $\phi_{\{2,4\}} = p_1 \wedge p_2 \wedge p_3$ and $\mathcal{E}_{\{2,4\}} = \emptyset$. Since $maxc_{\{1,3\}} = 2 > maxc$, we have $\Phi' = \{(\phi_{\{2,4\}}, \mathcal{E}_{\{2,4\}})\}$ and $maxc = 2$. We then handle $(\phi_{\{1,3\}}, \mathcal{E}_{\{1,3\}})$. We have $I = \{1\}$. So $\mathsf{MCS}(I) = \mathsf{CardM}(I) = \{I\}$ and $maxc_{\{1,3\}} = 1$. Since $max_{\{1,3\}} < maxc$, we do not need to consider dealing with $(\phi_{\{1,3\}}, \mathcal{E}_{\{1,3\}})$. Thus, we have $\Phi = \{(\phi_{\{2,4\}}, \mathcal{E}_{\{2,4\}})\}$. Since $\mathcal{E}_{\{2,4\}} = \emptyset$, the algorithm terminates and returns $B = p_1 \wedge p_2 \wedge p_3$, this coincides with the result in Example 1.

We show that Algorithm 1 results in the syntactical counterpart of our merging operator.

**Proposition 4** *Let $\mathcal{E} = \{B_1, ...., B_n\}$ be a possibilistic profile and $\mu$ a formula representing the integrity constraint. Suppose $B$ is the knowledge base obtained by Algorithm 1, then $B \equiv \Delta_\mu^{Lex}(\mathcal{E})$.*

## 5 Compatibility with Propositional Merging

In this section, we consider the compatibility of our merging operator with propositional merging by evaluating it with basic postulates (IC0)-(IC8) given in [11]. It is well-known that classical logic is a special case of possibilistic logic in which all the formulas have the same level of priority. That is, given a knowledge base $K = \{\phi_1, ..., \phi_n\}$, we can relate it to a possibilistic knowledge base $B_K = \{(\phi_1, 1), ..., (\phi_n, 1)\}$. Thus, our merging operator can be applied to merge propositional knowledge bases. We show that our merging operator is reduced to the $\Delta^{d_H, GMin}$ operator defined in [7], which is defined as follows. The *drastic distance* $d_H$ between two interpretations is defined as $d_H(\omega_1, \omega_2) = 0$ when $\omega_1 = \omega_2$ and 1 otherwise.

**Definition 7** *[7] Let $E = \{K_1, ..., K_n\}$ be a propositional knowledge profile and mu a formula. For any interpretation $\omega$, suppose $d_{d_H, Gmin}(\omega, E)$ is defined as the list of numbers $(d_1, ..., d_n)$ obtained by sorting in increasing order the set $\{d(\omega, K_i) \mid K_i \in E\}$. The models of $\Delta_\mu^{d_H, GMin}(E)$ are the models of $\mu$ that are minimal w.r.t. the lexicographic order induced by the natural order.*

**Proposition 5** *Let $E = \{K_1, ..., K_n\}$ be a propositional knowledge profile and $\mu$ a formula. Then $\Delta_\mu^{Lex}(E) = \Delta_\mu^{d_H, GMin}(E)$.*

Our merging operator is also related to the $\Delta^{C4}$ operator defined in [9], which selects the set of subsets of $E$ that are consistent with $\mu$ and are maximal with respect to cardinality. Let $E$ be a propositional knowledge profile and $\mu$ be a formula. Let $\mathsf{CardM}(E,\mu) = \{F \subseteq E \mid \bigwedge_{K_i \in F} K_i \wedge \mu \not\vdash \bot, \forall F' \subseteq E : |F'| > |F| \Rightarrow \bigwedge_{K_i \in F'} K_i \wedge \mu \vdash \bot\}$.

**Proposition 6** *Let $E$ be a propositional knowledge profile and $\mu$ be a formula. Then $\Delta^{Lex}_\mu(E) = \bigvee_{F \in \mathsf{CardM}(E,\mu)} (\bigwedge_{K_i \in F} K_i \wedge \mu)$.*

Proposition 6 shows that when each knowledge base is viewed as a formula, the $\Delta^{Lex}$ operator is equivalent to the $\Delta^{C4}$ operator.

Since the $\Delta^{d_H, GMin}$ operator satisfies all the basic postulates for propositional merging, our merging operator $\Delta^{Lex}$ also satisfies these postulates in the propositional case, i.e., when knowledge bases to be merged are propositional knowledge bases.

**Proposition 7** *$\Delta^{Lex}$ satisfies (IC0)-(IC8) given in [11] in the propositional case.*

Proposition 7 shows the advantage of our semantic merging operator over others as it has been shown in [17, 13] that existing semantic possibilistic merging operators do not satisfy all (IC0)-(IC8) in the propositional case. Another interesting result is that it is possible to encode the $\Delta^{C4}$ operator by using our merging operator $\Delta^{Lex}$ in case a knowledge base is considered as a set of formulas.

**Proposition 8** *Let $E = \{K_1, ..., K_n\}$ be a profile and $\mu$ be a formula. Suppose we split each knowledge base $K_i = \{\phi_1, ..., \phi_{n_i}\}$ into a set of bases $K'_1 = \{\phi_1\}, ..., K'_{n_i} = \{\phi_{n_i}\}$. If there are two duplicated knowledge bases after splitting, we only keep one of them. Let the resultant profile be $E' = \{K'_1, ..., K'_l\}$, then we have $\Delta^{Lex}_\mu(E') = \Delta^{C4}_\mu(E)$.*

## 6 Related Work

We compare our merging operator with existing ones given in the literature. Like our merging operator, the merging operators given in [4, 2] are semantically defined by aggregation of possibility distributions. The difference is that we apply the lexicographic aggregation function whilst the merging operators given in [4, 2] apply other aggregation operators, such as a *conjunctive operator* or a *disjunctive operator*. It has been shown in [17, 13] that merging operators given in [4, 2, 17] do not satisfy basic postulates for propositional merging given in [10] in the propositional case. The merging operators proposed in [5] are defined by aggregating weights of formulas inferred by possibilistic knowledge bases to be merged. The authors generalized postulates for propositional merging given in [10] by considering the logical closure of possibilistic inference and have shown that a merging operator satisfying all of the generalized postulates must be a *strictly monotonic* operator. Such a merging operator is similar to the conjunctive operator given in [2] and it thus suffers from the similar problems, that is, it suffers from the "drowning effect" and is not compatible with propositional merging. In contrast, our merging operator does not have such problems. One may argue that the disadvantage of our merging operator is that the weighted information is lost in the resulting knowledge base after merging. However, as we have discussed in the Introduction, this is not an issue if we want to know if a formula can be inferred from the merged knowledge base. Let us consider an example. Let $\mu = p_1 \vee p_2$, $\mathcal{E} = \{B_1, B_2\}$, where $B_1 = \{(\neg p_2, 0.8), (p_4, 0.6)\}$ and $B_2 = \{(p_2, 0.9), (p_1, 0.8), (p_3, 0.6)\}$. By applying the product merging operator given in [2], we obtain the result of merging as $B \equiv_s \{(\mu, 1)\} \cup B_1 \cup B_2 \cup \{(p_1 \vee p_2, 0.99), (p_1 \vee p_3, 0.96), (\neg p_2 \vee p_3, 0.88), (p_2 \vee p_4, 0.96), (p_3 \vee p_4, 0.76)\}$. The inconsistency degree of $B$ is 0.8. Therefore, $p_1$ and $p_3$ cannot be inferred from $B$ by using the possibilistic inference due to the drowning effect. However, applying our merging operator, we get $\Delta^{Lex}_\mu(\mathcal{E}) = \mu \wedge B_2^*$. It is easy to check that $\Delta^{Lex}_\mu(\mathcal{E}) \vdash B^*_{>0.8}$, and $p_1$ and $p_3$ can be inferred from $\Delta^{Lex}_\mu(\mathcal{E})$. This example shows that there are cases where the formulas inferred from the result of our merging operator by applying classical inference can be more than those inferred from the result of the product merging operator by applying possibilistic inference.

The adaptive merging operator given in [12] has no semantic definition and has not been evaluated by logical properties. The merging approaches given in [3] are dependent on the syntactic form of formulas in a possibilistic knowledge base. The merging operator given in [14] is defined by a classical merging operator, thus it is compatible with propositional merging. However, it is only evaluated by four postulates that generalize some basic postulates given in [11] and its rationality w.r.t. postulates is still not fully recognized. The merging operator proposed in [15] is also based on lexicographic ordering. However, it is applied to merging prioritized knowledge bases that do not share a common scale, thus is different from ours.

## 7 Conclusion and Discussion

In this paper, we have considered the problem of merging knowledge bases in possibilistic logic. We first pre-sented a novel merging operator in possibilistic logic based on the lexicographic ordering. *The key difference between our merging operator and existing ones is that we do not require that the weighted information be kept in the merged knowledge base.* We proposed some new logical postulates for possibilistic merging and showed that our merging operator satisfies all of them. We then provided an algorithm for computing the syntactical counterpart of our merging operator. Finally, we analyzed the compatibility of our merging operator with propositional merging. We showed that our merging operator can be reduced to the operator $\Delta^{d_H,GMin}$ defined in [7] operator in the propositional case and it satisfies postulates ($IC$0-$IC$8) given in [11]. We also showed that it is possible to encode the $\Delta^{C4}$ operator by using our merging operator $\Delta^{Lex}$ in case a knowledge base is considered as a set of formulas.

We have generalized some logical postulates for propositional merging operators, as a future work, we will try to provide a representation theorem for the generalized postulates. We will also investigate other merging operators in possibilistic logic that map a possibilistic profile to a propositional knowledge base.

## Acknowledgments

Guilin Qi is partially supported by Excellent Youth Scholars Program of Southeast University under grant 4009001011 and National Science Foundation of China under grant 60903010. We would like to thanks three anonymous reviewers for their helpful comments.